\documentclass{article}

\PassOptionsToPackage{numbers, compress}{natbib}

\usepackage[preprint]{neurips_2024}

\usepackage[utf8]{inputenc} 
\usepackage[T1]{fontenc}    
\usepackage{hyperref}       
\usepackage{url}            
\usepackage{booktabs}       
\usepackage{amsfonts}       
\usepackage{nicefrac}       
\usepackage{microtype}      
\usepackage{xcolor}         
\usepackage{graphicx}
\usepackage{algorithmic}
\usepackage{algorithm}
\usepackage{ascmac}

\title{Development and bilingual evaluation of Japanese medical large language model within reasonably low computational resources}

\author{%
  Issey Sukeda \\
  Department of Mathematical Informatics, Graduate School of Information Science and Technology,\\
  The University of Tokyo,\\
  Tokyo, Japan\\
  \texttt{sukeda-issei006@g.ecc.u-tokyo.ac.jp} \\
}

\begin{document}

\maketitle

\begin{abstract}
The recent success of large language models (LLMs) and the scaling law has led to a widespread adoption of larger models. Particularly in the healthcare industry, there is an increasing demand for locally operated LLMs due to security concerns. However, the majority of high quality open-source LLMs have a size of 70B parameters, imposing significant financial burdens on users for GPU preparation and operation. To overcome these issues, we present a medical adaptation based on the recent 7B models, which enables the operation in low computational resources. 
We compare the performance on medical question-answering benchmarks in two languages (Japanese and English), demonstrating that its scores reach parity with or surpass those of currently existing medical LLMs that are ten times larger. We find that fine-tuning an English-centric base model on Japanese medical dataset improves the score in both language, supporting the effect of cross-lingual knowledge transfer. We hope that this study will alleviate financial challenges, serving as a stepping stone for clinical institutions to practically utilize LLMs locally. 
Our evaluation code is available at \url{https://github.com/stardust-coder/japanese-lm-med-harness}.
\end{abstract}

\section{Introduction}

In recent years, while the development of LLMs in the medical field has been progressing, there still remains a significant gap between their development and practical application in clinical settings. One of these gaps is the operational environment of LLMs. 
Current medical LLMs can broadly be divided into two types: the large models developed in closed fashion by big tech companies~\cite{singhal2023large,singhal2023towards,tu2024towards,nori2023medprompt} and the open-source models with fewer parameters. The former is said to have anywhere from tens of billions to trillions of parameters or more and is not freely accessible, or is typically accessible only via API services. This raises security concerns for clinical institutions dealing with patient's personal information, causing hesitation in implementing them in real clinical settings. On the other hand, to improve customizability and accessibility, many other medical LLMs have been released, most of which are based on the LLama series~\cite{touvron2023llama,touvron2023llama2}, a series of open-source LLMs developed by Meta Inc. The number of parameters in these model developments falls into two main categories: around 7 $\sim$ 8B and 70B. Generally, larger models outperform smaller ones, a phenomenon known as the scaling law~\cite{kaplan2020scaling}. However, smaller models are computationally more efficient in pretraining, fine-tuning, and inference.
For each clinical institute, it is challenging to allocate sufficient computational resources due to budget constraints and other factors. To enable the practical use of medical LLMs, achieving substantial performance with smaller LLMs like 7 $\sim$ 8B models that can operate within each institute's realistic computational environments is essential. 

Since many existing medical LLMs are English-centric, it is believed that there is a strong push in non-English-speaking countries to develop similar medical LLMs in the native languages of users, such as patients and doctors, from a practical standpoint, as this would be more user-friendly.
This is done by integrating medical domain adaptation and language adaptation. Particularly in Japan, although several studies have evaluated the capabilities of the commercial GPT models in handling medical queries~\cite{jpn-med-exam_gpt4}, the number of local model developments lags behind compared to English-centric models~\cite{sukeda2023jmedlora,sukeda2024jmedlora}. Details about the related works is deferred to Appendix~\ref{appendix:related-works}.

The aim of this study is to verify whether we can avoid operating LLMs as large as 70B-parameter or more; i.e., we aim at achieving enhancement or extraction of Japanese medical intelligence only using limited computational resources, thereby enabling the resulting model to be deployed across numerous clinical institutes in Japan.
We conduct the evaluation based on accuracy using the only existing Japanese medical benchmark IgakuQA~\cite{jpn-med-exam_gpt4}, which is essentially the NMLE (National Medical Licensing Examination in Japan). However, instead of aiming to develop models specifically to pass the medical licensing exam, our focus here is to achieve substantial performance with a 7B medical LLM.

\section{Method}

We follow the standard way of developing domain-specialized LLMs. 
First, we prepare the medical corpus from a medical journal and conducted the full-parameter training. We call this process MFPT (Medical Full-Parameter Training) in this work. 
Subsequently, we conduct the LoRA (Low Rank Adaptation)~\cite{hu2021lora} fine-tuning using question-answering (Q\&A) dataset. In this work, we refer to this process as MPEFT (Medical Parameter-Efficient Fine-Tuning). 

\subsection{Models}

For the English-centric base model, Llama3~\cite{dubey2024llama3herdmodels} or Qwen2~\cite{qwen2} have recently been two choices due to their superior performances (see e.g. \cite{open-llm-leaderboard-v2}). 
We build upon the Qwen2~\cite{qwen2} as the base model architecture for medical adaptation since Qwen2 generally performs better than Llama3 in solving medical benchmark tasks as in Table~\ref{tab:performance-ja} and Table~\ref{tab:performance-en}.

We specifically focus on the 7B-parameter model of this series, which enables us to perform full-parameter training even with limited computational resources. Moreover, the Qwen2 series~\cite{qwen2} is released under the Apache-2.0 license, which provides significant advantages for broader use.

\subsection{Fine-tuning}
\subsubsection{Medical Full-Parameter Training (MFPT)}

First, we conduct the full-parameter training using our own medical corpus, following the established approach by previous researches suggesting that continual-pretraining or additional training before instruction-tuning is effective~\cite{singhal2023large}. Specifically, we used the \textit{naika-text} corpus, which is composed of 6120 lines of Japanese sentences (3.5M letters) extracted from the Journal of the Japanese Society of Internal Medicine. The model after MFPT for 5 epochs is referred to as \textit{Ours-MFPT} in Table~\ref{tab:performance}.

\subsubsection{Medical Parameter-Efficient Fine-Tuning (MPEFT)}

Following MFPT, MPEFT is performed using the training split of the USMLE (United States Medical Licensing Examination), which includes 10178 training examples in a Q\&A format. Since our goal is to develop a Japanese LLM, this data --- originally in English --- is translated into Japanese by a medical doctor manually\footnote{This translated dataset will not be made public.}. On the other hand, we also use the original English USMLE for comparative experiments. Note that we do not include any data in training dataset from IgakuQA because the IgakuQA dataset is not large\footnote{Otherwise, to avoid data leakage, we need to split the dataset into training and evaluation sets, further reducing their sizes.}.

In this procedure, we apply LoRA (Low Rank Adaptation)~\cite{hu2021lora}, a parameter-efficient fine-tuning method that can drastically save computational resources (especially GPU memory) without significant performance loss compared to full-parameter training. 
To compare each contribution of MFPT and MPEFT, we apply it to both the base Qwen2-7B-Instruct and \textit{Ours-MFPT} model, respectively. \textit{Ours-MFPT} after MPEFT for 5 epochs is referred to as \textit{Ours-MPEFT} in Table~\ref{tab:performance}.








\subsection{A unified evaluation method}

In medical LLM research, numerous studies have reported benchmark scores derived from question-answering tasks. These benchmarks facilitate comparative analysis. However, even when the test datasets are identical across studies, variations in the experimental settings surrounding LLM inference often preclude truly equitable comparisons. In our paper, we report unified evaluation scores measured by our own experiments, instead of quoting those from previous studies.
To facilitate the evaluation method presented in this paper and in future works, and to further encourage the development of medical LLMs, we will make our evaluation codes, which can be executed with a single line of script, publicly available with customization options.

\subsubsection{Benchmark dataset for evaluation}

We curate four bilingual medical benchmarks in Japanese and English to assess model performance and language tendencies. IgakuQA features five-choice questions, while the other benchmarks use four-choice questions.

\paragraph{IgakuQA~\cite{jpn-med-exam_gpt4}}  is constructed based on the national medical license exam from 2018 to 2022 in Japan. Both the original Japanese dataset and the English-translated dataset are released at \url{https://github.com/jungokasai/IgakuQA}, including 1450 five-choice questions and answers.

\paragraph{MedQA~\cite{jin2020disease}}  is composed of the medical license exam in the US, USMLE for short. We only include its evaluation split with 1273 samples in our benchmark. Since the original dataset is in English, the Japanese-translated version was prepared by hand.

\paragraph{MedMCQA~\cite{pmlr-v174-pal22a}}  (Multi-Subject Multi-Choice Dataset for Medical domain) is a four-choice question-answering task designed to address real-world medical entrance exam questions. 
We only include its evaluation split with 4183 samples in our benchmark. Since the original dataset is in English, the Japanese-translated version was prepared by hand.

\paragraph{MMLU~\cite{hendrycks2020measuring}} (Massive Multi-task Language Understanding) is a four-choice Q\&A dataset including 57 tasks covering various subjects. We extract five medical-related subjects, i.e., anatomy (135 samples), clinical knowledge (265 samples), college medicine (173 samples), medical genetics (100 samples), and professional medicine (272 samples).

\paragraph{JMMLU~\cite{yin2022should}} is a Japanese-translation of a subset of MMLU, recently prepared as a counterpart of MMLU in our own language. Our evaluation covers anatomy (132 samples), clinical knowledge (150 samples), college medicine (151 samples), medical genetics (99 samples), and professional medicine (150 samples).


\subsubsection{Task and evaluation metric}

Experimental settings in the inference side generally include prompting, metric, and the hyperparameter of the generation process.

\paragraph{Prompting} Prompting mainly consists of the following three factors, which are not consistent at all in the previous reports: (i) prompt template (ii) the number of few-shot examples (iii) other algorithmic prompting techniques. 
For the template, while Alpaca~\cite{alpaca} has been the defacto standard, MedPaLM-2~\cite{singhal2023towards} and Meditron~\cite{chen2023meditron70b} are evaluated with a slightly different prompt template, respectively. 
The number of few-shot examples is still controversial. Basically, the larger the better but when the prompt becomes too long, the model tends to ignore the former instructions.
In addition, to improve the benchmark scores, Chain-of-thought (CoT) prompting~\cite{wei2022chain} is commonly used, followed by self-consistency~\cite{wang2022self} in MedPaLM~\cite{singhal2023large}, and ensemble refinement in MedPaLM-2~\cite{singhal2023towards}. In our experiments, the standard CoT prompt is applied; see Appendix~\ref{appendix:prompt} for the detailed prompting strategy.

\paragraph{Metric} In multiple-choices Q\&A tasks, evaluations based on accuracy is common, i.e., we let the model pick one choice as its response and compare it with the correct answer. 
The typical method involves having models select each answer from five labels [``a'', ``b'', ``c'', ``d'', ``e''] and verifying by exact match --- an approach also used by Kasai et al.~\cite{jpn-med-exam_gpt4}--- we instead instruct the model to output the words or sentences from the alternatives directly, as the labels themselves do not provide essential information in practice.

To calculate accuracy, \textit{Exact match} has been the most objective and the easiest metric. Also, \textit{Gestalt accuracy} was proposed as an alternative by Sukeda et al.~\cite{sukeda2023jmedlora,sukeda2024jmedlora}, which is a more robust metric to admit a slight mistake for LLMs. In our experiments, we employ the \textit{Gestalt accuracy}.

\paragraph{Hyperparameter of the generation process} Commonly, the trainer and the inference pipeline of LLMs is implemented by huggingface transformers~\cite{wolf2019huggingface}, which requires to specify several hyperparameters for text generation, e.g., the sampling method, the number of beam for beam search, the temperature for sampling, repetition penalty, and so on. 
Practically, high temperature and high repetition penalty along with sophisticated sampling methods are recommended to achieve natural and various text generation. On the other hand, in many studies, deterministic results are more preferable for reproducability and thus beam search with one beam is typically employed but not in every case. In our following experiments, we employ the deterministic setting.

\subsubsection{Miscellaneous}
Although several ad-hoc techniques to improve LLMs' performance including few-shot prompting~\cite{brown2020language}, mixture of experts/agents~\cite{wang2024mixture}, and model merging~\cite{akiba2024evolutionary}, have been developed recently, we do not employ these techniques in our main study, as we expect their application can be independently dissociated from the core potential of the LLM that we aim to examine. In other words, these techniques can be readily integrated to one another in practical use cases.

\section{Results}

\subsection{Our resulting model}

Table~\ref{tab:performance} lists the five types we created, along with their training data and the accuracy of IgakuQA in both English and Japanese. We performed fine-tuning of the model with five different settings by varying the training steps.
The Ours-LoRA models surpass MFPT (fine-tuning with low-rank adaptation) on the base Qwen2 model~\cite{qwen2} in both English and Japanese. In contrast, Ours-MFPT involved full parameter tuning as described in Section 2.2.1. Additionally, Ours-MPEFT, which is based on Ours-MFPT, further improves upon MPEFT in the same manner.

From the accuracy of IgakuQA, it is observed that both the MFPT and MPEFT processes have steadily contributed to score improvement. Specifically, for Japanese IgakuQA, Ours-MPEFT(ja) model achieved a 10.8\% increase in accuracy compared to the base model, while it improved by 2.5\% in English, though the improvement was smaller. The score of this model exceeds 50\% accuracy in both English and Japanese, exhibiting substantial bilingual performance as a 7B model. Hereafter, we will refer to our best model Ours-MPEFT(ja) as JMedLLM-v1-7B.

\begin{table}[t]
    \centering
    \scalebox{0.7}{
    \begin{tabular}{cccccc|cc|c}
         model& base model & size&license& MFPT& MPEFT & en(\%)&ja(\%)&Ave.\\ \hline\hline
          Ours-LoRA(en) &Qwen2&7B&CC-BY-NC-SA-4.0&-&USMLE(en)&47.7&41.5&44.6\\
          Ours-LoRA(ja) &Qwen2&7B&CC-BY-NC-SA-4.0&-&USMLE(ja)&51.1&48.6&49.8 \\
          Ours-MFPT&Qwen2&7B&CC-BY-NC-SA-4.0&naika-text(ja)&-&47.3&46.6& 46.9 \\
          Ours-MPEFT(en) &Qwen2&7B&CC-BY-NC-SA-4.0&naika-text(ja)&USMLE(en)&46.2&44.2&45.2\\
          Ours-MPEFT(ja)&Qwen2&7B&CC-BY-NC-SA-4.0&naika-text(ja)&USMLE(ja)&50.6&52.3&\bf{51.4} \\
    \end{tabular}
    }
    \caption{Benchmark accuracy of our models evaluated with IgakuQA in English(en) and Japanese(ja).} 
    \label{tab:performance}
\end{table}

\subsection{Comparison with other open-source LLMs}

\paragraph{Setup}
All generation is performed in zero-shot, meaning no example input-output pairs are provided to the LLM. The 7B models are used as-is, while all 70B models are loaded using 4-bit quantization techniques by default to conserve computational resources. We did not conduct multiple runs or observe deviations; instead, we employed deterministic inference without sampling.

\paragraph{Models}

Our baseline includes all of the major medical LLMs from previous works. Each of these models are developed by continual training or fine-tuning on each base model, mostly Llama series. For English-centric models, Meditron~\cite{chen2023meditron70b} and Med42~\cite{christophe2024med42} are Llama2-based while OpenBioLLM~\cite{OpenBioLLMs} and Med42-v2~\cite{christophe2024med42-v2} is Llama3-based. For Japanese-centric models, Preferred-Llama3-MedSwallow-70B is a recent Llama3-Swallow-based medical LLM reported to achieve better accuracy than GPT-4\footnote{\url{https://openai.com/index/gpt-4/}} in solving NMLE under their experimental settings\footnote{\url{https://tech.preferred.jp/ja/blog/llama3-preferred-medswallow-70b/}}. Different from the Llama2 series, the Llama3 series is empirically known to have substantial multilingual ability~\cite{dubey2024llama3herdmodels}, thus we also evaluate both English and Japanese performance for the models derivedd from the Llama3 series.
In addition, we add to our baselines the base models for general purpose, which are used as the backbone of each medical LLM. Specifically, Llama3~\cite{dubey2024llama3herdmodels} is added as the English-centric baseline while Llama3-Swallow~\cite{llama3swallow} is added as the Japanese-centric baseline. Moreover, for the 7$\sim$8B scale, two Japanese-centric models --- Youko~\cite{sawada2024release} and Llama-3-ELYZA~\cite{elyzallama2024} --- are added for comparison.
In our experiments, we utilize the instruct version whenever available. 
The link to each specific model is listed in Appendix~\ref{appendix:instruction-models}.


\paragraph{Results}

In Table~\ref{tab:performance-ja}, we observe that at the 7$\sim$8B scale, JMedLLM-v1 outperforms other baselines including base models and medical LLMs on average in four Japanese medical benchmarks by more than $10\%$. It is notable that even at the 70B scale, JMedLLM-v1 outperforms other baselines except Preferred-Llama3-MedSwallow, surpassing 50\% accuracy in IgakuQA, MedQA, and MedMCQA. Specifically, JMedLLM-v1 outperforms 70B-parameter OpenBioLLM by as much as $7.5\%$.
One of the possible reasons of the improved performance is that JMedLLM-v1 is based on Qwen2 as its backbone, which achieves superior performance to Llama3 series with the same size. However, in Table~\ref{tab:70b-dif-ja}, it is shown that additional training on medical dataset signifcantly contribute to the score improvement despite the discrepancy between training data and evaluation benchmarks in general. In fact, except the cases of solving Japanese medical benchmarks with OpenBioLLM, the among four models show substantial improvement.
Especially, JMedLLM-v1 is further improved and outperforms the base Qwen2 by 13.9\%. 

On the other hand, in Table~\ref{tab:performance-en}, we observe that JMedLLM-v1 outperforms other models of similar size on average across four English medical benchmarks, being the only model to surpass 50\% accuracy. 
Despite being 10 times larger, the existing English-centric medical LLMs, Meditron~\cite{chen2023meditron70b} and Med42~\cite{christophe2024med42}, do not outperform JMedLLM-v1.
However, two 70B-parameter models, OpenBioLLM and Preferred-Llama3-MedSwallow, achieve higher scores than JMedLLM-v1, surpassing $60\%$ in Gestalt accuracy on average. Table~\ref{tab:70b-dif-en} further exhibits the score improvement of each LLM, where we can see JMedLLM-v1 has improved by $13.3\%$ in Gestalt accuracy on average also in English medical benchmarks.

Overall, Preferred-Llama3-MedSwallow scores the highest among Japanese medical models, followed by our JMedLLM-v1. OpenBioLLM~\cite{OpenBioLLMs} performs best in English medical tasks but performs worse in Japanese. However, Preferred-Llama3-MedSwallow and JMedLLM-v1 also show strong bilingual performance. Among 7B parameter models, our model stands out as the best performer .

\renewcommand{\arraystretch}{1.3}
\begin{table}[t]
    \centering
    \scalebox{0.7}{
    \begin{tabular}{l|cc|cccc|c}
        model(-size) & base & language & IgakuQA(ja) & MedQA(ja) & MedMCQA(ja) & JMMLU & Ave.(ja) \\ \hline\hline
        Llama3-70B\cite{llama3swallow} & Llama3 & en & 43.1 & \bf{40.9} & 37.2 & 45.3 & \bf{41.6} \\ 
        Llama3-Swallow-70B\cite{llama3swallow} & Llama3 & ja & 44.6 & 32.9 & 33.7 & 37.5 & 37.2 \\ 
         OpenBioLLM-70B\cite{OpenBioLLMs} & Llama3 & en & 35.6 & 35.4 & \bf{39.9} & \bf{54.6} & 41.4 \\ 
         Preferred-Llama3-MedSwallow-70B & Llama3 & ja & \bf{62.6} & \bf{55.6} & \bf{43.4} & \bf{58.4}& \bf{55.0} \\  \hline
        Llama3-8B\cite{dubey2024llama3herdmodels} & Llama3 & en & 23.8 & 28.7 & 31.7 & 30.8 & 28.8 \\ 
        Youko-8B\cite{sawada2024release} & Llama3 & ja & 33.5 & 31.1 & 34.0 & 35.7 & 33.6 \\ 
        Llama3-Swallow-8B\cite{llama3swallow} & Llama3 & ja & 28.5 & 26.9 & 31.3 & 30.4 & 29.3 \\ 
        Llama-3-ELYZA-JP-8B\cite{elyzallama2024} & Llama3 & ja & 38.0 & 33.4 & 34.1 & 42.9 & 37.1 \\ 
         MMedLlama3-8B\cite{qiu2024towards} & Llama3 & en & 31.5 & 34.3 & 33.9 & 40.1 & 35.0 \\ 
        Qwen2-7B\cite{qwen2} & Qwen2 & en & \bf{44.6} & 30.8 & 31.5 & 33.2 & 35.0 \\ 
         JMedLLM-v1-7B (Ours) & Qwen2 & ja & \bf{52.3} & \bf{51.2} & \bf{41.2} & \bf{50.8} & \bf{48.9} \\ 
    \end{tabular}
    }
    \vspace{5mm}
    \caption{\textbf{JMedLLM-v1 against open-source baselines in Japanese medical benchmarks.} This table shows the main
results of JMedLLM-v1's medical task performance in Japanese against other best-performing open-source medical LLMs measured by the Gestalt accuracy(\%). Top 3 scores in each row are marked in bold.}
    \label{tab:performance-ja}
\end{table}

\begin{table}[t]
    \centering
    \scalebox{0.76}{
    \begin{tabular}{ccc|cccc}
         base LLM&$\to$&Japanese LLM&IgakuQA& MedQA& MedMCQA&JMMLU \\ \hline
         Llama3-70B~\cite{dubey2024llama3herdmodels} &$\to$&Llama3-Swallow-70B~\cite{llama3swallow} & $+1.5$ & $-8.0$& $-3.5$ &$-7.8$ \\
         Llama3-8B~\cite{dubey2024llama3herdmodels} &$\to$&Youko-8B~\cite{sawada2024release}& $+9.3$ & $+2.4$ & $+2.3$ & $+4.9$\\ 
         Llama3-8B~\cite{dubey2024llama3herdmodels} &$\to$&Llama3-Swallow-8B~\cite{llama3swallow} & $+4.7$ & $-1.8$& $-0.4$ & $-0.4$\\
         Llama3-8B~\cite{dubey2024llama3herdmodels} &$\to$&Llama-3-ELYZA-JP-8B\cite{elyzallama2024} & $+14.2$ & $+4.7$&  $+2.4$ & $+12.1$\\     
    \end{tabular}
    }
    \vspace{5mm}
    
    \scalebox{0.76}{
    \begin{tabular}{ccc|cccc}
         base LLM&$\to$& medical LLM&IgakuQA& MedQA& MedMCQA&JMMLU \\ \hline
         Llama3-70B~\cite{dubey2024llama3herdmodels} &$\to$& OpenBioLLM-70B~\cite{OpenBioLLMs}& $-7.5$ & $-5.5$& $+2.7$ & $+9.3$\\
         Llama3-Swallow-70B~\cite{llama3swallow} &$\to$&Preferred-Llama3-MedSwallow-70B&$+18.0$& $+22.7$& $+9.7$ & $+20.9$\\ 
         Llama3-8B~\cite{dubey2024llama3herdmodels} &$\to$&MMedLlama3-8B\cite{qiu2024towards} & $+7.7$ & $+5.6$& $+2.2$& $+9.3$\\     
         Qwen2-7B\cite{qwen2}  &$\to$&JMedLLM-v1-7B (Ours) & $+7.7$ & $+20.4$& $+9.7$ & $+17.6$\\
    \end{tabular}
    }
    \vspace{5mm}
    \caption{\textbf{Score improvements of LLMs compared to each base model in Japanese benchmarks.} This table shows the difference in scores between each medical LLM and its corresponding base model, as presented in Table 2.}
    \label{tab:70b-dif-ja}
\end{table}

\begin{table}[t]
    \centering
    \scalebox{0.7}{
    \begin{tabular}{l|cc|cccc|c}
        model(-size) & base & language & IgakuQA(en) & MedQA(en) & MedMCQA(en) & MMLU & Ave.(en) \\ \hline\hline
         Meditron-70B~\cite{chen2023meditron70b}&	Llama2&	en	&29.9&	44.7&	32.8&	49.6&	39.3 \\
         Med42-70B~\cite{christophe2024med42}&	Llama2&	en	&45.0&	56.2&	\bf{48.2}&	60.9& 52.6 \\
        Llama3-70B\cite{dubey2024llama3herdmodels} & Llama3 & en & 38.3 & \bf{57.7} & 38.8 & \bf{63.7} & 49.6 \\ 
        Llama3-Swallow-70B\cite{llama3swallow} & Llama3 & ja & \bf{52.8} & 39.0 & 43.0 & 51.2 & 46.5 \\ 
         OpenBioLLM-70B\cite{OpenBioLLMs} & Llama3 & en & \bf{58.5} & \bf{70.2} & \bf{65.0} & \bf{80.0} & \bf{68.4} \\ 
         Preferred-Llama3-MedSwallow-70B & Llama3 & ja & \bf{55.0} & \bf{61.3} & \bf{52.9} & \bf{68.1} & \bf{59.3} \\ \hline
        Llama3-8B\cite{dubey2024llama3herdmodels} & Llama3 & en & 35.0 & 43.0 & 39.1 & 41.3 & 39.6 \\ 
        Youko-8B\cite{sawada2024release} & Llama3 & ja & 38.1 & 34.1 & 29.4 & 44.6 & 36.6 \\ 
        Llama3-Swallow-8B\cite{llama3swallow} & Llama3 & ja & 34.4 & 30.8 & 36.0 & 38.8 & 35.0 \\ 
        Llama-3-ELYZA-JP-8B\cite{elyzallama2024} & Llama3 & ja & 20.7 & 40.6 & 37.3 & 44.7 & 35.8 \\ 
         MMedLlama3-8B\cite{qiu2024towards} & Llama3 & en & 26.4 & 36.8 & 37.5 & 37.7 & 34.6 \\ 
        Qwen2-7B\cite{qwen2} & Qwen2 & en & 46.4 & 36.9 & 34.7 & 43.1 & 40.3 \\ 
         \bf{JMedLLM-v1-7B (Ours)} & Qwen2 & ja & 50.6 & 54.6 & 46.1 & 63.0 & \bf{53.6} \\ 
    \end{tabular}
    }
    \vspace{5mm}
    \caption{\textbf{JMedLLM-v1 against open-source baselines in English  medical benchmarks.} This table shows the main
results of JMedLLM-v1's medical task performance in English against other best-performing open-source medical LLMs measured by the Gestalt accuracy(\%). Top 3 scores in each row are marked in bold.}
    \label{tab:performance-en}
\end{table}

\begin{table}[t]
    \centering
    \scalebox{0.76}{
    \begin{tabular}{ccc|cccc}
         base LLM&$\to$&Japanese LLM&IgakuQA& MedQA& MedMCQA&JMMLU \\ \hline
         Llama3-70B~\cite{dubey2024llama3herdmodels} &$\to$&Llama3-Swallow-70B~\cite{llama3swallow} &  $+14.5$ & $-18.7$& $+4.2$ & $-12.5$\\
         Llama3-8B~\cite{dubey2024llama3herdmodels} &$\to$&Youko-8B~\cite{sawada2024release}& $+3.1$&$-8.9$ & $-9.7$&$+3.3$ \\ 
         Llama3-8B~\cite{dubey2024llama3herdmodels} &$\to$&Llama3-Swallow-8B~\cite{llama3swallow} & $-0.6$ & $-12.2$& $-2.9$ & $-2.5$ \\
         Llama3-8B~\cite{dubey2024llama3herdmodels} &$\to$&Llama-3-ELYZA-JP-8B\cite{elyzallama2024} & $-14.3$ & $-2.4$& $-1.8$ & $+3.4$\\     
    \end{tabular}
    }
    \vspace{5mm}
    
    \scalebox{0.76}{
    \begin{tabular}{ccc|cccc}
         base LLM&$\to$& medical LLM&IgakuQA& MedQA& MedMCQA&JMMLU \\ \hline
         Llama3-70B~\cite{dubey2024llama3herdmodels} &$\to$& OpenBioLLM-70B~\cite{OpenBioLLMs}& $+20.2$ & $+12.5$& $+26.2$ & $+16.3$\\
         Llama3-Swallow-70B~\cite{llama3swallow} &$\to$&Preferred-Llama3-MedSwallow-70B&$+2.2$&$+22.3$ & $+9.9$& $+16.9$ \\ 
         Llama3-8B~\cite{dubey2024llama3herdmodels} &$\to$&MMedLlama3-8B\cite{qiu2024towards} & $-8.6$ & $-6.2$& $-1.6$ & $-3.6$ \\     
         Qwen2-7B\cite{qwen2}  &$\to$&JMedLLM-v1-7B (Ours) & $+4.2$ & $+17.7$& $+11.4$ & $+19.9$\\
    \end{tabular}
    }
    \vspace{5mm}
    \caption{\textbf{Score improvements of LLMs compared to each base model in English benchmarks.} This table shows the difference in scores between each medical LLM and its corresponding base model, as presented in Table 4.}
    \label{tab:70b-dif-en}
\end{table}

\subsection{Comparison with the state-of-the-art}

Table~\ref{tab:sota} shows the gap between the top-3 open-source models from Table~\ref{tab:performance-en} and three closed models: Med-Gemini~\cite{saab2024capabilitiesgeminimodelsmedicine}, Med-PaLM2~\cite{singhal2023towards}, and GPT-4. Although these score comparisons are for reference only, as the evaluation settings are not completely aligned and each score is taken from previous reports, the closed models generally outperform the open-source models. Med-Gemini-L 1.0~\cite{saab2024capabilitiesgeminimodelsmedicine} is outstanding in MedQA, achieving 91.1\% accuracy. Meanwhile, OpenBioLLM~\cite{OpenBioLLMs}, with 70B parameters, approaches GPT-4, particularly in the MedMCQA scores.
\begin{table}[t]
    \centering
    \begin{tabular}{c|cccc}
         Model name&IgakuQA(ja) & MedQA(en)  & MedMCQA(en)&  MMLU \\ \hline\hline
         Med-Gemini-L 1.0&  - & 91.1$^{(a)}$ & - & -\\
         Med-PaLM2&-&85.4$^{(b)}$&72.3$^{(b)}$& 88.4$^{(b)}$\\
         GPT-4&  78.2$^{(c)}$ & 78.8$^{(d)}$  & 69.5$^{(d)}$    & 86.0$^{(d)}$  \\ \hline
         {\small Preferred-Llama3-MedSwallow}&62.6 &61.2  & 52.9 & 68.1 \\
         OpenBioLLM& 35.6  &70.2& 65.0 & 80.0\\
         JMedLLM-v1&  52.3 & 54.6 &46.1&63.0 \\
    \end{tabular}
    \caption{\textbf{JMedLLM-v1 against the state-of-the-art LLMs.} This table shows the scores cited from previous studies. Note that each score differs in inference settings. ${(a)}$ Cited from \cite{saab2024capabilitiesgeminimodelsmedicine}. ${(b)}$ : Cited or calculated from \cite{singhal2023towards}, where the ensemble refinement method is used, which is computationally costly. ${(c)}$ : Calculated based on \cite{jpn-med-exam_gpt4}, where 3-shot in-context learning is used. ${(d)}$ Cited or calculated from zero-shot performances in \cite{nori2023capabilities}.}
    \label{tab:sota}
\end{table}

\renewcommand{\arraystretch}{1}

\subsection{Low computational resources}

Our MFPT phase took only 7.5 hours on 8 NVIDIA A100 GPUs. Our MPEFT phase took only 28.5 hours on 4 NVIDIA V100 GPUs. All experiments were conducted using ABCI, a Japanese domestic cloud computing infrastructure. These computational burdens are significantly smaller than Meditron (332 hours on 128 A100 GPUs for training) and Med42 (the Condor Galaxy 1 supercomputer for full-parameter fine-tuning). \footnote{For example, Amazon Web Services (\url{https://aws.amazon.com/?nc1=h_ls}) provides these GPU instances as cloud environment. The cost for training can be simulated as $32.77 \times 7.5 + 12.24 \times 28.5 = 595$ USD.}

On the other hand, all the evaluation experiments for 70B models with quantization were run on 4 NVIDIA V100 GPUs, whereas for 7B models were run on 1 NVIDIA V100 GPUs.

\section{Discussion}

\paragraph{Model performance and size from a practical use perspective} 

Since LLMs generally perform better as their size increases, a phenomenon known as the scaling law~\cite{kaplan2020scaling}, it is inevitable to confront this tradeoff in practice.
The aim of our work is to develop 7B-parameter medical LLMs to the fullest extent possible, so that clinical institutes do not necessarily have to rely on external API services or the computationally burdensome 70B models. 

In our work, we demonstrate that the 7B-parameter model, with a solid base model and fine-tuning process using a domain-specific corpus, can potentially outperform 70B models in a medical question-answering benchmarks both in Japanese and English.
Although its performances shown in Table~\ref{tab:performance-ja} and \ref{tab:performance-en} are not totally sufficient, 
this result highlights the potential for the practical use of medical LLMs in clinical institutions, as 7B-parameter models can operate relatively quickly with modest resources, such as a single GPU in a standard environment.

To implement medical LLMs in real clinical institutions, how far should we make progress? One way to set the performance necessary for practical use is to compare it with large, closed-source LLMs such as GPT-4, the MedPaLM series, and the recent top-performing Med-Gemini. Table~\ref{tab:sota} shows that open-source medical LLMs still have a significant gap compared to closed-source models in benchmark scores. However, for example, when compared to Med-PaLM2, JMedLLM-v1 with 7B parameters achieves from 60 to 70\% of the performance with only less than 2.5\% parameters. \footnote{Med-PaLM2 is a closed medical LLM based on PaLM2, which is also a closed model. According to unofficial news sources (\url{https://www.cnbc.com/2023/05/16/googles-palm-2-uses-nearly-five-times-more-text-data-than-predecessor.html}, accessed 2024/8/10), PaLM2 is reported to have 340B parameters based on internal documents. Official information is not yet available.}


\paragraph{Score improvements resulting from fine-tuning in different languages}

By comparing Tables~\ref{tab:70b-dif-ja} and \ref{tab:70b-dif-en}, it is observed that Japanese adaptation tends to result in only a small improvement in scores or even a decline in performance. In contrast, medical adaptation generally leads to more significant score improvements across medical benchmarks in both languages.

Individually, the fine-tuning of MMedLlama3~\cite{qiu2024towards} in a multilingual context enhances performance in Japanese while causing degration in English. Conversely, OpenBioLLM is English-centric, leading to significant improvements in English benchmarks but sometimes causing a decline in performance in certain Japanese benchmarks.
On the other hand, both Preferred-Llama3-MedSwallow and JMedLLM-v1 are finetuned with Japanese dataset, yet they improve performance in both languages. 

A significant improvement in Table~\ref{tab:70b-dif-ja} for JMedLLM-v1 on MedQA and for Preferred-Llama3-MedSwallow on IgakuQA can be attributed to the alignment between their evaluation benchmarks and training datasets, although the details for the latter are not fully disclosed.\footnote{According to a tech blog by the developers of Preferred-Llama3-MedSwallow, the fine-tuning dataset includes past NMLE data up to 2017. URL: \url{https://tech.preferred.jp/ja/blog/llama3-preferred-medswallow-70b/}} However, the significant improvement in MMLU/JMMLU is impressive, whereas IgakuQA and MedMCQA show more variable results, which may be related to task difficulty: MMLU/JMMLU covers high school to university levels, MedMCQA targets graduate school level, and IgakuQA and MedQA focus on national exam level.

\paragraph{Knowledge extraction by fine-tuning} 
As mentioned in the previous paragraph, it is well-known that alignment between fine-tuning and evaluation tasks generally leads to significant improvements in model performance. However, our experiments indicate that improvements can occur even in the absence of such alignment. Although USMLE and NMLE (or other benchmark tasks) are situated within the same medical domain, their questions and answer choices are not perfectly identical. The observed enhancements in our MPEFT model suggest that the base model already possesses a foundational level of medical knowledge. Rather than introducing new medical knowledge through fine-tuning with USMLE data, this process appears to activate latent capabilities within the model. These findings imply that training on tasks that are similar but not identical to the target task can still contribute to improved model performance.  To further investigate or quantify this phenomenon, it would be necessary to identify the specific information required for accurately answering a specific target question, which currently appears to be technically infeasible.

\paragraph{Cross-lingual knowledge transfer} 

When adapting English-centric LLMs to local languages, is there a trade-off in performance for English while learning a new language? In our English-translated IgakuQA evaluations, as shown in Table~\ref{tab:performance}, an unexpected observation is that our models demonstrated improvements of +1.4\% for MFPT and +2.9\% for MPEFT, despite the fine-tuning training data consisting solely of Japanese texts. This phenomenon is also evident in Table~\ref{tab:70b-dif-en} for JMedLLM-v1 and Preferred-Llama3-MedSwallow. Although these improvements are smaller compared to those observed in Japanese benchmarks listed in Table~\ref{tab:70b-dif-ja}, they are still noteworthy. A similar phenomenon is observed in our comparative LoRA experiments without the MFPT step, where accuracy improved by 1.8\% in English and by 5.2\% in Japanese compared to the base model (Qwen2-Instruct~\cite{qwen2}). Overall, our results suggest that cross-lingual training on English-centric models can be effective not only for acquiring local languages but also for enhancing their performance in English.

\section{Limitations}

\subsection{Insufficient data resources}

The amount and variation of medical corpus have been insufficient for training LLMs, particularly in Japanese. In our study, we utilized medical examinations from the US as a data resource; however, this may introduce a risk of reflecting biases inherent to US medicine. Not only should the selection and preparation of the training dataset be further improved, but bias correction methods across different countries, cultures, and contexts also need to be studied further to ensure practical use.

\subsection{Exploration on evaluation method}

\paragraph{Question on multiple-choice question-answering as benchmarks}
This work does not explore the validity of the evaluation method in depth; instead, we prioritize unification. However, in the study of LLMs, the development of a fair evaluation method is eagerly anticipated.
Evaluating the performance of medical LLMs with question-answering task, which is often based on the medical licensing exam, is questioned~\cite{ness2024medfuzz} in terms of the risk  reproducing social biases in clinical decision making.

\paragraph{Mismatch prompt formatting in training and evaluation} 

Some LLMs are designed to adhere to specific prompt formatting, especially when instruction-tuning~\cite{wei2021finetuned} is involved. Empirically, 70B models are sufficiently generalized and capable of handling variations in prompt formatting, whereas 7B models tend to perform worse in this regard. Nonetheless, LLMs are expected to perform optimally when the prompt format during inference is specified correctly. For instance, Meditron~\cite{chen2023meditron70b} follows the ChatML format~\cite{chatml}, whereas our prompting strategies differ significantly (see Appendix~\ref{appendix:prompt}). This discrepancy may contribute to the poorer performance of Meditron observed in Table~\ref{tab:performance-en}.

\paragraph{Tokenizer specification} 

We also point out that our models employ the \textit{Byte-Pair Encoding tokenization}~\cite{wang2020neural} as well as the backbone Qwen series. The use of it for LLMs has been argued~\cite{bostrom2020byte}, and may not be optimal especially for Japanese LLMs. 

\subsection{Data contamination}
Typically, the amount of training corpus for base models is extensive and not entirely publicly available. Therefore, although many reports assert that the fine-tuning process does not explicitly allow for contamination, it is not possible to guarantee that the evaluation datasets used for benchmarking (such as IgakuQA, MedQA, MedMCQA, and MMLU/JMMLU in our case) are free from contamination. Once the contamination occurs, it causes data leakage, which artificially inflates benchmark scores. For instance, Sukeda et al.\cite{sukeda2024-70b} demonstrate in an ablation study that fine-tuning an LLM on USMLE and then evaluating it with the same data can easily lead to accuracy surpassing 80\%. Therefore, it should be noted that a significant leap in accuracy shown in Table\ref{tab:70b-dif-ja} or \ref{tab:70b-dif-en} might be at risk of being caused by data leakage unless the data usage is clearly specified.

\subsection{Model quantization}

We regret that we used 4-bit quantization for evaluating all the 70B-parameter models due to limited computational resources. While quantization is generally known to speed up inference and potentially degrade performance, it is sometimes argued that 4-bit quantization can still maintain good performance in practice. Quantitatively evaluating the degradation caused by quantization will be a focus of future work.

\section{Conclusion}

In this work, we develop the leading  7B medical LLM and demonstrate that it achieves performance comparable to or better than existing 4-bit-quantized 70B-parameter medical LLMs on Japanese medical Q\&A benchmarks. Moreover, our model also performs well on English counterparts even without additional training on English data. Our 7B model can be trained and operated in environments with limited GPU resources, addressing financial and security concerns for clinical institutes seeking to adopt practical, medical-specific LLMs.

\section{Acknowledgments and Disclosure of Funding}

The Japanese-translated version of USMLE was provided by Dr. Hisahiko Sato (not made public).
The Japanese-translated version of MedQA and MedMCQA were provided by Mr.Junfeng Jiang (\url{https://huggingface.co/datasets/Coldog2333/JMedBench}). 
We also thank the developer of SWIFT~\cite{swift}, which our implementation for training models is based on, and Dr. Jun Sese and Dr. Shinnosuke Sawano for helpful comments.
This work was supported by AIST KAKUSEI project (FY2023).

\bibliographystyle{plain}
\bibliography{merge}

\appendix

\section{Related works} \label{appendix:related-works}

The relevant models and their training dataset are curated in Table~\ref{tab:llm-dict} as a reference.

\subsection{Before LLMs}
The Japanese language model has been developed following precedents set by English research, particularly in the medical domain. Recently, Japanese medical language model research was ignited by UTH-BERT~\cite{kawazoe2021clinical}, which was developed by pretraining BERT~\cite{devlin2018bert} with approximately 120M clinical texts stored at UTokyo Hospital as the first medical language model in Japanese. Afterwards, JMedRoBERTa~\cite{sugimoto2023jmedroberta} was developed based on RoBERTa model using the abstract and main text of the non-public medical papers. 

\subsection{Medical LLMs in English}

In open source community, two lines of family have been developed: the Llama family~\cite{touvron2023llama,touvron2023llama2} and the Mistral family~\cite{jiang2023mistral,jiang2024mixtral}. Generally, the Llama family tends to be a single model while the Mistral family has evolved in the direction of the Mixture-of-Experts. Specifically in biomedical areas, several medical LLMs have been built upon 70B-parameter Llama2 or Llama3~\cite{sukeda2023jmedlora,chen2023meditron70b,christophe2024med42,OpenBioLLMs,cheng2024adapting}, while 7B-parameter Biomistral~\cite{labrak2024biomistral} and BiMediX~\cite{pieri2024bimedix} have been derived from Mistral 7B~\cite{jiang2023mistral} and Mixtral-8x7B~\cite{jiang2024mixtral}, respectively.

\subsection{Medical LLMs in Japanese}

Compared to English-centric models, the Japanese medical LLMs lack its number. Instead of developing the model from scratch, these models are developed based on powerful English-centric models. 
The first attempt in this domain was JMedLoRA~\cite{sukeda2023jmedlora}, which conducted the QLoRA~\cite{dettmers2023qlora} instruction-tuning on Llama2-70B~\cite{singhal2023towards}. After the Japanese general LLM named Swallow~\cite{fujii2024continual} was released, Sukeda et al.~\cite{sukeda2024-70b} performed the similar fine-tuning on Llama2~\cite{singhal2023towards}, Xwin~\cite{xwin-lm}, and Swallow~\cite{fujii2024continual}, suggesting the potential of Japanese base model to be improved largely in medical question-answering by instruction-tuning. Furthermore, after the release of Llama3-Swallow~\cite{llama3swallow}, which is based on Llama3~\cite{dubey2024llama3herdmodels}, Preferred Network Inc. performed the continual training via QLoRA~\cite{dettmers2023qlora} using their non-public medical training data, which is released as Preferred-Llama3-MedSwallow.

\begin{table}
    \centering
    \scalebox{0.85}{
    \begin{tabular}{ccc}
        Model name & Training dataset & \#Data \\ \hline \hline
        Llama2~\cite{touvron2023llama2} &   See~\cite{touvron2023llama2}.& 2T tokens\\\hline
        Meditron~\cite{chen2023meditron70b} & \begin{tabular}{c}
             Clinical Guidelines,  \\
             PubMed Abstracts,\\
             PubMed Papers,\\
             Experience Replay
        \end{tabular}&\begin{tabular}{c}
             0.107B tokens\\
             5.48B tokens\\
             40.7B tokens\\
             0.420B tokens
        \end{tabular}\\\hline
        Med42~\cite{christophe2024med42} &   See \cite{christophe2024med42}.& \begin{tabular}{c}
        411064 medical samples\\
        295649 general domain samples
        \end{tabular}
        \\\hline
        Swallow~\cite{fujii2024continual} & \begin{tabular}{c}
        Swallow corpus,\\
        Japanese Wikipedia,\\
        the RefinedWeb\\
        The Pile
        \end{tabular}& \begin{tabular}{c}
             312.1B characters\\
             \\
             104.9B tokens\\
             \\
        \end{tabular}\\ \hline
        MedSwallow~\cite{sukeda2024-70b} &  Japanese-translated USMLE & 12723 samples\\\hline
        Llama3~\cite{dubey2024llama3herdmodels} & \begin{tabular}{c} Web-curated multilingual data \\covering 176 languages\end{tabular} &15T tokens\\ \hline
        Youko~\cite{qiu2024towards} & \begin{tabular}{c}
        Japanese CC-100,\\
        Japanese C4, \\
        Japanese OSCAR,\\
        The Pile, Wikipedia, \\
        rinna curated Japanese dataset 
        \end{tabular}& 22B tokens\\\hline
        Qwen2~\cite{qwen2} & \begin{tabular}{c}Multilingual data\\ supporting 30 languages\end{tabular} & 7T tokens\\ \hline
        OpenBioLLM~\cite{OpenBioLLMs}&\begin{tabular}{c}
         Custom Medical Instruct dataset\\
         DPO dataset
         \end{tabular}& unknown\\ \hline
        Preferred-Llama3-MedSwallow-70B& own medical corpus & unknown\\ \hline
        JMedLLM-v1~ (Ours) & naika dataset, USMLE & 3.5M characters + 10178 samples \\ \hline
    \end{tabular}
    }
    \caption{Training dataset of existing LLMs. The number of tokens is presented for each dataset if available. Otherwise, the number of samples is presented.}
    \label{tab:llm-dict}
\end{table}

\subsection{Medical benchmarks}

To develop the domain-specific LLMs, the evaluation benchmarks are of great importance. Here, we review the existing medical evaluation benchmarks here.

\paragraph{MultiMedBench~\cite{tu2024towards}} is an open source multimodal medical benchmark developed for assessing the multimodal medical model named MedPaLM-M, including three of the MultiMedQA tasks used
to evaluate MedPaLM~\cite{singhal2023towards}, and radiology report summarization.

\paragraph{MIRAGE~\cite{xiong2024benchmarking}} is a benchmark for medical LLMs and retrieval augmented generation(RAG), which includes MedQA~\cite{jin2020disease}, MedMCQA~\cite{pmlr-v174-pal22a}, PubMedQA~\cite{jin2019pubmedqa}, MMLU Subsets (Medicine)~\cite{hendrycks2020measuring} and BioASQ-QA~\cite{krithara2023bioasq}.

\paragraph{The Open Medical-LLM Leaderboard~\cite{Medical-LLMLeaderboard}} is a standarized platform that provides a setup specifically designed for the medical domain, which includes MedQA~\cite{jin2020disease}, MedMCQA~\cite{pmlr-v174-pal22a}, PubMedQA~\cite{jin2019pubmedqa}, and MMLU Subsets (Medicine and Biology)~\cite{hendrycks2020measuring}.

\paragraph{CMB~\cite{wang2023cmb}} is a comprehensive medical benchmark in Chinese, which comprises multiple-choice questions from qualification exams, and complex clinical diagnostic questions derived from real case studies.

\paragraph{MMedBench~\cite{qiu2024towards}} is a comprehensive multilingual medical benchmark including six languages, English, Japanese, Chinese, French, Spanish, and Russian. 

\section{Instruct models} \label{appendix:instruction-models}

For base models in our experiments, we utilize the instruct version whenever available. Specifically for non-medical base models, we use 
Meta-Llama-3-70B-Instruct\footnote{\url{https://huggingface.co/meta-llama/Meta-Llama-3-70B-Instruct}}, 
Llama-3-Swallow-70B-Instruct-v0.1\footnote{\url{https://huggingface.co/tokyotech-llm/Llama-3-Swallow-70B-Instruct-v0.1}},
Meta-Llama-3-8B-Instruct\footnote{\url{https://huggingface.co/meta-llama/Meta-Llama-3-8B-Instruct}}, 
llama-3-youko-8b\footnote{\url{https://huggingface.co/rinna/llama-3-youko-8b}},
and
Qwen2-7B-Instruct\footnote{\url{https://huggingface.co/Qwen/Qwen2-7B-Instruct}}.

For medical models, we use
Meditron-70B\footnote{\url{https://huggingface.co/epfl-llm/meditron-70b}}, 
Med42-70B\footnote{\url{https://huggingface.co/m42-health/med42-70b}},
Llama3-OpenBioLLM-70B\footnote{\url{https://huggingface.co/aaditya/Llama3-OpenBioLLM-70B}}, 
MMedLlama3-8B\footnote{\url{https://huggingface.co/Henrychur/MMed-Llama-3-8B}},
and
Llama3-Preferred-MedSwallow-70B\footnote{\url{https://huggingface.co/pfnet/Llama3-Preferred-MedSwallow-70B}}.


\section{Prompting strategies in inference} \label{appendix:prompt}

For the prompt template, we follow the Chain-of-Thought (CoT) prompt~\cite{wei2022chain} used in Med-PaLM 2~\cite{singhal2023towards} by default. Japanese templates are prepared as well through translation by ChatGPT\footnote{\url{https://chatgpt.com/}}.
To let the model solve the given question-answering tasks, the question sentence is input into \{\{instruction\}\} and four or five candidate choices are input into \{\{input\}\}.

\paragraph{Chain-of-Thought (CoT) prompt}
\begin{quote}
\#\#\# Instruction:\\
The following are multiple choice questions about medical knowledge. Solve them in a step-by-step fashion, starting by summarizing the available information. Output a single option from the five options as the final answer.\\
\#\#\# Input:\\
\{\{instruction\}\}\\
\{\{input\}\}\\
\#\#\# Response:\\
\end{quote}
\noindent Although prompt template selection tends to be ad-hoc since we cannot know which one is better than the other in advance, we choose this one as our experimental setting because it seemed to be slightly superior in Sukeda et al.\cite{sukeda2024-70b}. Few-shot inference has known to be effective strategy, however we do not apply it as our standard experimental setting since the number of shots is always arbitrary. Moreover, providing more examples in a prompt tends to lead to better performance; however, it entails a token limit issue.
Therefore, we align the experimental settings in each run by evaluating zero-shot performance.

As comparative studies, we additionally evaluate our best model with different prompting strategies. First we attempt the following standard prompt:
\paragraph{Alpaca~\cite{alpaca} prompt}
\begin{quote}
Below is an instruction that describes a task, paired with an input that provides further context. Write a response that appropriately completes the request.\\
\#\#\# Instruction:\\
\{\{instruction\}\}\\
\#\#\# Input:\\
\{\{input\}\}\\
\#\#\# Response:
\end{quote}
\noindent As a result, Table~\ref{tab:alpaca} shows that using the Alpaca prompt template led to slightly worse performances both in English and Japanese. However, in theory, there should be no superiority or inferiority since the meaning of given instruction is almost identical. In experiments by Sukeda et al.~\cite{sukeda2023jmedlora}, the superiority of these two types of prompts reversed depending on the experimental settings, making it difficult to conclude. Hence, a difference in accuracy of a few percentage points may be considered negligible.

Subsequently, we observe a few-shot performance using the following prompt:
\paragraph{Few-shot inference with CoT}
\begin{quote}
    \#\#\# Instruction:\\
    The following are multiple choice questions about medical knowledge. Solve them in a step-by-step fashion, starting by summarizing the available information. Output a single option from the five options as the final answer.\\
    \#\#\# Input:\\
    Which of the following is not a mandatory explanation to be provided to participants in human genome\/gene analysis research?\\
    The purpose of the research, The freedom to consent, Methods for anonymity, Disadvantages of participation, Assurance of research results sharing\\
    \#\#\# Response:\\
    Assurance of research results sharing\\
    \#\#\# Input:\\
    A 57-year-old man lost consciousness and collapsed while working to remove sludge from a manhole at a sewage treatment plant. A colleague who entered to assist also suddenly lost consciousness and collapsed. Which of the following is the most likely cause? Select two.\\
    Oxygen deficiency, Hydrogen sulfide poisoning, Carbon monoxide poisoning, Carbon dioxide poisoning, Nitrogen dioxide poisoning\\
    \#\#\# Response:\\
    Oxygen deficiency, Hydrogen sulfide poisoning\\
    \#\#\# Input:\\
    A 28-year-old woman at 30 weeks of gestation has a fundal height of 22 cm and almost no amniotic fluid is detected on abdominal ultrasound examination. What is the most likely condition in the fetus?\\
    Esophageal atresia, Ventricular septal defect, Renal hypoplasia, Anorectal malformation, Fetal hydrops\\
    \#\#\# Response:\\
    Renal hypoplasia\\
    \#\#\# Input:\\
    \{\{instruction\}\}\\
    \{\{input\}\}\\
    \#\#\# Response:
\end{quote}
\noindent In 1-shot experiments, only the first example was included, while the whole prompt was applied for 3-shot experiments.
As shown in Table~\ref{tab:few-shot}, few-shot prompting technique did not contribute to the score improvement. While few-shot examples are believed to instruct the model via in-context learning, our model already has the ability to follow the instruction to choose one option from five alternatives. Thus, the given few-shot examples may function as noisy information unrelated to the target question.

\begin{table}[t]
  \begin{minipage}[t]{.45\textwidth}
    \begin{center}
    \vspace{-6mm}
      \begin{tabular}{c|cc}
            Template & en(\%)  & ja(\%) \\ \hline
            CoT & \bf{50.2} & \bf{52.5} \\
            Alpaca & 49.7  & 49.7  \\
        \end{tabular}
        \caption{Differences by prompt templates}
        \label{tab:alpaca}
    \end{center}
  \end{minipage}
  \hfill
  \begin{minipage}[t]{.45\textwidth}

    \begin{center}
      \begin{tabular}{c|cc}
         & en(\%) & ja(\%) \\ \hline
        0-shot with CoT & \bf{50.2} & \bf{52.5}\\
        1-shot with CoT & 45.8 & 45.7\\
        3-shot with CoT & 46.7 & 47.8\\
    \end{tabular}
    \caption{Difference by the number of few-shot examples}
    \label{tab:few-shot}
    \end{center}
  \end{minipage}
\end{table}

\end{document}